\documentclass[runningheads]{llncs}
\usepackage[T1]{fontenc}
\usepackage{graphicx}
\usepackage{bm}
\usepackage{bibentry}
\usepackage{amsmath}
\usepackage{xcolor}
\usepackage{tabularx}
\usepackage{booktabs} 
\usepackage[table]{xcolor}
\usepackage{caption}
\usepackage{subcaption}
\usepackage{mathtools}
\usepackage{amsfonts}
\usepackage[linesnumbered,ruled,vlined,procnumbered]{algorithm2e}%me for pseudo-code
\begin{document}
\title{DS-Span: Single-Phase Discriminative Subgraph Mining for Efficient Graph Embeddings}
\titlerunning{\textbf{DS-Span}}
\author{Yeamin Kaiser\inst{1} \and Muhammed Tasnim Bin Anwar\inst{1} \and Bholanath Das\inst{1}}
\authorrunning{Kaiser et al.}
\institute{University of Dhaka}

\maketitle              % typeset the header of the contribution
\begin{abstract}
Graph representation learning seeks to transform complex, high-dimensional graph structures into compact vector spaces that preserve both topology and semantics. Among the various strategies, subgraph-based methods provide an interpretable bridge between symbolic pattern discovery and continuous embedding learning. Yet, existing frequent or discriminative subgraph mining approaches often suffer from redundant multi-phase pipelines, high computational cost, and weak coupling between mined structures and their discriminative relevance. We propose \textbf{DS-Span}, a single-phase discriminative subgraph mining framework that unifies pattern growth, pruning, and supervision-driven scoring within one traversal of the search space. DS-Span introduces a \emph{coverage-capped eligibility mechanism} that dynamically limits exploration once a graph is sufficiently represented, and an \emph{information-gain-guided selection} that promotes subgraphs with strong class-separating ability while minimizing redundancy. The resulting subgraph set serves as an efficient, interpretable basis for downstream graph embedding and classification. Extensive experiments across benchmarks demonstrate that DS-Span generates more compact and discriminative subgraph features than prior multi-stage methods, achieving higher or comparable accuracy with significantly reduced runtime. These results highlight the potential of unified, single-phase discriminative mining as a foundation for scalable and interpretable graph representation learning.

\end{abstract}
\section{Introduction}

Graphs provide a natural and powerful formalism for representing relational data in diverse domains such as chemistry, bioinformatics, transportation, and social networks. In these settings, the ability to learn compact yet expressive representations of graphs is crucial for tasks such as classification, similarity search, and clustering. Graph representation learning aims to transform discrete graph structures into continuous vector spaces while preserving structural and semantic information. 

Among the various approaches to graph representation, \emph{subgraph-based methods} occupy a unique position by bridging symbolic and continuous perspectives. Instead of directly parameterizing the graph through neural architectures, these methods first discover \emph{frequent} or \emph{discriminative} substructures that recur across the dataset and then use their presence or embedding to form interpretable graph-level features. This paradigm provides both transparency and efficiency, especially when domain interpretability and explainability are essential.

Classical algorithms such as \textbf{gSpan}~\cite{gspan} introduced depth-first search (DFS)-based canonical coding and pattern growth for efficient enumeration of frequent subgraphs. Later methods extended this idea toward discriminative mining, identifying subgraphs that maximize class-separating power rather than sheer frequency. Despite their success, most existing discriminative mining frameworks share three key limitations:

\begin{enumerate}
    \item \textbf{Redundant multi-phase mining:} Frequent and discriminative phases are typically decoupled, leading to repeated subgraph enumeration and redundant computation.
    \item \textbf{Weak coupling between supervision and mining:} Class-label information is often used only after enumeration, rather than guiding the mining process itself.
    \item \textbf{Limited scalability:} Exhaustive enumeration on large or dense graphs produces substantial overlap among candidate subgraphs, causing combinatorial explosion and long runtimes.
\end{enumerate}

To address these challenges, we propose \textbf{DS-Span}, a \emph{single-phase discriminative subgraph mining} framework that integrates subgraph enumeration, pruning, and discriminative scoring into a unified search process. DS-Span maintains the canonical DFS-based traversal of gSpan but introduces several novel mechanisms to enhance both efficiency and discriminative relevance:
\begin{itemize}
    \item \textbf{Coverage-capped eligibility:} graphs are dynamically excluded from further exploration once their feature coverage exceeds a user-defined threshold, preventing redundant pattern growth.
    \item \textbf{Information-gain-guided search:} candidate subgraphs are scored using information gain to prioritize those that best separate graph classes while minimizing overlap among features.
    \item \textbf{Single-phase pruning:} all constraints on support, coverage, and discriminative power are applied concurrently within one traversal, eliminating the need for iterative re-mining or post-hoc filtering.
\end{itemize}

By embedding these mechanisms directly into the mining process, DS-Span discovers a compact and highly discriminative subset of subgraphs that can serve as efficient inputs for downstream embedding and classification models. This unified framework drastically reduces redundancy and computational overhead, leading to substantial runtime savings without sacrificing predictive performance.

Our experimental study across diverse datasets demonstrates that DS-Span produces more informative and less redundant subgraph features than existing multi-phase methods. The resulting embeddings exhibit stronger class separation and lower variance across folds, underscoring the efficacy of single-phase, discriminatively guided subgraph mining as a foundation for scalable and interpretable graph representation learning.

\section{Proposed Approach}
\label{sec:method}

We present \textbf{DS-Span}, a single-phase discriminative subgraph mining framework that produces a compact, high-utility feature set for downstream graph embedding and classification. DS-Span preserves the DFS-code canonicalization and rightmost-path extensions of \textbf{gSpan}~\cite{gspan}, but differs in three design choices:
(i) a \emph{coverage-capped eligibility} mechanism that prunes graphs from further exploration once sufficiently represented,
(ii) a \emph{single-phase} mining schedule (no multi-phase restarts), and
(iii) a \emph{discriminative, coverage-constrained} feature selector.
These choices eliminate redundant re-mining, guide the search toward class-separating patterns, and improve time/memory efficiency.

\subsection{Preliminaries and Notation}
\label{subsec:prelim}

Let $\mathcal{D}=\{G_i\}_{i=1}^N$ be a collection of labeled graphs with (training) class labels $y_i \in \mathcal{C}$.
A graph $G=(V,E,\ell_V,\ell_E)$ has node labels $\ell_V:V\!\to\!\Sigma_V$ and edge labels $\ell_E:E\!\to\!\Sigma_E$.
For two graphs $S$ and $G$, we write $S \subseteq_{\text{iso}} G$ if $S$ is subgraph-isomorphic to $G$.
The \emph{support} of a pattern $S$ in $\mathcal{D}$ is
\[
\mathrm{supp}(S) \coloneqq \big|\{\, i \in [N] \,:\, S \subseteq_{\text{iso}} G_i \,\}\big|.
\]
We use DFS-code tuples $(u,v,\ell_u,\ell_v,\ell_e)$ and rightmost-path extensions as in \cite{gspan}.
A DFS code is \emph{canonical} if it is minimal in the gSpan lexicographic order; the canonical minimality test prevents duplicate patterns.

\paragraph{Coverage.}
For a set of mined subgraphs $\mathcal{F}$ and a graph $G_i$, define the coverage
$\mathrm{cov}(i;\mathcal{F}) \coloneqq
\big|\{\, S\in \mathcal{F}: S \subseteq_{\text{iso}} G_i \,\}\big|$.
Two thresholds govern coverage: a per-graph \emph{minimum coverage} $\textit{min\_cov}\in\mathbb{N}$,
and a per-graph \emph{cap} $\textit{cap} \coloneqq \gamma \cdot \textit{min\_cov}$ with $\gamma\!\ge\!1$.
Coverage influences \emph{eligibility}: graphs that reach the cap stop contributing to further extensions.

\paragraph{Discriminative score.}
For a candidate $S$, let $\mathcal{I}(S)=\{i: S \subseteq_{\text{iso}} G_i\}$ and $\overline{\mathcal{I}}(S)$ its complement.
Let $H(\cdot)$ denote Shannon entropy over class labels.
We use information gain
\[
IG(S) \coloneqq H(y) - \Big(\tfrac{|\mathcal{I}(S)|}{N}\, H(y\,|\,i\!\in\!\mathcal{I}(S)) \;+\; \tfrac{|\overline{\mathcal{I}}(S)|}{N}\, H(y\,|\,i\!\in\!\overline{\mathcal{I}}(S))\Big),
\]
and equivalently refer to \emph{weighted entropy} $WE(S) \coloneqq H(y)-IG(S)$ when minimizing.
A dataset-level \emph{coverage constraint} requires the selected feature set $\mathcal{F}_\star$ to cover at least a fraction $\tau$ of graphs:
$\big|\bigcup_{S\in \mathcal{F}_\star}\mathcal{I}(S)\big| \ge \tau N$.

\subsection{Single-Phase Prospective Mining with Coverage-Capped Eligibility}
\label{subsec:mining}

Classical discriminative pipelines often alternate between (i) frequent enumeration under a support threshold and (ii) post-hoc discriminative filtering, sometimes across multiple mining \emph{phases}. In contrast, DS-Span performs a \emph{single} DFS traversal that (a) enforces a global support threshold and (b) dynamically shrinks the set of graphs that are eligible to produce further extensions as their coverage grows.

Let $\delta\in(0,1]$ be the relative support threshold; we require $\mathrm{supp}(S)\ge \lceil \delta \cdot N\rceil$.
We maintain a per-graph coverage counter and an \emph{eligibility mask} $\mathcal{E}\subseteq [N]$ of graphs whose coverage is $<\textit{cap}$.
Rightmost-path extensions are collected \emph{only} from graphs in $\mathcal{E}$.
Whenever a new frequent, canonical extension $S'$ is accepted, we increment coverage for graphs in $\mathcal{I}(S')$ and remove any graph $i$ with $\mathrm{cov}(i)\ge \textit{cap}$ from $\mathcal{E}$, preventing redundant growth from already well-represented graphs.

\begin{algorithm}[t]
\small
\SetAlgoLined
\DontPrintSemicolon
\SetKwInOut{Input}{Input}
\SetKwInOut{Output}{Output}
\SetKwProg{Fn}{Function}{}{}
\SetKwFunction{Mine}{Mine}
\Input{$\mathcal{D}$: graphs; $\delta$: support; $\textit{min\_cov}$; $\gamma$; DFS-code canonical order}
\Output{$\mathcal{C}$: candidate subgraph set; $\mathrm{cov}$: per-graph coverage}
Initialize $\mathcal{C}\leftarrow\emptyset$;\; Initialize $\mathrm{cov}[i]\leftarrow 0$ and eligibility set $\mathcal{E}\leftarrow [N]$;\;
\Fn{\textsc{Mine}$(\mathrm{code},\mathcal{E})$}{
  $X \leftarrow$ \textsc{RightMostExtensions}$(\mathrm{code}, \{G_i:i\in \mathcal{E}\})$;\;
  \ForEach{$x \in X$}{
    $\mathrm{code}' \leftarrow \mathrm{code}\,\cup\,\{x\}$;\;
    \If{\textsc{IsCanonical}$(\mathrm{code}')$ and $\mathrm{supp}(\mathrm{code}') \ge \lceil \delta N\rceil$}{
      $\mathcal{C}\leftarrow \mathcal{C} \cup \{\mathrm{code}'\}$;\;
      \ForEach{$i \in \mathcal{I}(\mathrm{code}')$}{
        $\mathrm{cov}[i] \leftarrow \mathrm{cov}[i] + 1$;\;
        \If{$\mathrm{cov}[i] \ge \gamma\cdot \textit{min\_cov}$}{ $\mathcal{E}\leftarrow \mathcal{E}\setminus\{i\}$ }
      }
      \textsc{Mine}$(\mathrm{code}',\mathcal{E})$\;
    }
  }
}
\textsc{Mine}$(\emptyset,\mathcal{E})$;\;
\caption{Single-Phase Mining with Coverage-Capped Eligibility}
\label{alg:mine}
\end{algorithm}

\paragraph{Coverage completion (fairness top-up).}
A single traversal can still leave some graphs with $\mathrm{cov}(i) < \textit{min\_cov}$.
To avoid under-representing such graphs in downstream learning, we \emph{complete} coverage by enumerating the smallest canonical codes \emph{within each under-covered graph} and adding them to $\mathcal{C}$ until $\textit{min\_cov}$ is met for that graph.
These \emph{coverage fillers} have low support by construction and are subsequently down-weighted or removed by the discriminative selector (\ref{subsec:filtering}).

Unlike multi-phase methods that restart mining with relaxed thresholds, DS-Span grows patterns once and prunes \emph{graphs} (eligibility) as soon as they are sufficiently represented. This reduces extension generation, avoids revisiting already saturated regions, and cuts redundant isomorphism checks, all while maintaining the anti-monotonicity guarantees of support and the duplicate-avoidance guarantees of DFS-code minimality.

\subsection{Discriminative Selection under a Coverage Constraint}
\label{subsec:filtering}

The mined candidate pool $\mathcal{C}$ is distilled into a compact feature set $\mathcal{F}_\star$ by optimizing for discriminative utility subject to dataset coverage.
We formulate selection as a coverage-constrained set function optimization:

\begin{equation}
\label{eq:selection}
\max_{\mathcal{F}\subseteq \mathcal{C},\,|\mathcal{F}|\le K}\;\;\sum_{S\in \mathcal{F}} IG(S)
\quad \text{s.t.}\quad 
\Big|\bigcup_{S\in \mathcal{F}}\mathcal{I}(S)\Big| \;\ge\; \tau N,
\end{equation}

where $K$ is an optional feature budget and $\tau\in(0,1]$ controls dataset coverage (e.g., $\tau=0.95$).
We use a simple greedy strategy that is effective and easy to reproduce: sort candidates by $IG(S)$ descending, add a candidate if it \emph{improves} the coverage of under-covered graphs and respects the budget, and stop when the coverage constraint is met.

\begin{algorithm}[t]
\small
\SetAlgoLined
\DontPrintSemicolon
\SetKwInOut{Input}{Input}
\SetKwInOut{Output}{Output}
\Input{$\mathcal{C}$: candidates; $\{y_i\}$: labels; $K$; $\tau$}
\Output{$\mathcal{F}_\star$: selected features}
Compute $IG(S)$ for all $S\in \mathcal{C}$;\; Sort $\mathcal{C}$ by $IG(S)$ descending;\;
$\mathcal{F}_\star\leftarrow\emptyset$, $\mathcal{U}\leftarrow\emptyset$ \tcp*{$\mathcal{U}$: covered graph indices}
\ForEach{$S\in \mathcal{C}$}{
  \If{$|\mathcal{F}_\star|<K$ \textbf{ and } $\big|\mathcal{U}\cup \mathcal{I}(S)\big| > |\mathcal{U}|$}{
    $\mathcal{F}_\star\leftarrow \mathcal{F}_\star\cup\{S\}$;\;
    $\mathcal{U}\leftarrow \mathcal{U}\cup \mathcal{I}(S)$;\;
    \If{$|\mathcal{U}| \ge \tau N$}{ \textbf{break} }
  }
}
\caption{Coverage-Constrained Discriminative Selection}
\label{alg:filter}
\end{algorithm}

Post-hoc filters that ignore coverage tend to pick many near-duplicates of the same motif, hurting generalization and wasting capacity. Our coverage constraint explicitly promotes \emph{representative} features, reducing redundancy. Empirically (Sec.~\ref{experiments}), this increases accuracy with \emph{fewer} features and reduces variance across folds.

\subsection{Feature Embedding and Classification}
\label{subsec:embedding}

From the selected discriminative subgraphs $\mathcal{F}_\star = \{S_k\}_{k=1}^{K'}$, each graph $G_i$ is represented by a normalized binary incidence vector
\[
(x_i)_k =
\begin{cases}
\frac{1}{|\{\,j : S_j \subseteq_{\text{iso}} G_i\,\}|}, & \text{if } S_k \subseteq_{\text{iso}} G_i,\\[4pt]
0, & \text{otherwise.}
\end{cases}
\]
This normalization divides by the number of subgraphs present in $G_i$, ensuring that larger graphs do not produce disproportionately high feature magnitudes. The resulting feature matrix $X = [x_1, x_2, \ldots, x_N]^\top$ is sparse and interpretable, directly encoding the presence of mined substructures.

\vspace{0.5em}
\noindent\textbf{Embedding model.}
Following the architecture of \textbf{DisFPGC}~\cite{mta-bp}, we employ a shallow CBOW-style embedding network that learns two parameter matrices $W \in \mathbb{R}^{K'\times E}$ and $W' \in \mathbb{R}^{E\times N}$, where $E$ denotes the embedding dimension and $N$ the number of graphs. Each forward pass computes hidden activations
\[
h_i = W^\top x_i, \qquad
u_i = {W'}^\top h_i,
\]
and outputs $\hat{y}_i = \mathrm{softmax}(u_i)$.
The model is trained to minimize cross-entropy loss between $\hat{y}_i$ and the true one-hot label vector $t_i$:
\[
\mathcal{L} = -\sum_{i=1}^{N} t_i^\top \log \hat{y}_i.
\]
The same optimization procedure and criteria described in DisFPGC are used here. Consequently, any difference in classification performance arises solely from DS-Span’s feature quality.

\subsection{Design Contrasts with Prior Literature}
\label{subsec:contrast}

\noindent\textbf{Versus gSpan \cite{gspan}.}
Both use DFS-code canonicalization and rightmost expansion.
DS-Span adds \emph{coverage-capped eligibility} and a \emph{discriminative, coverage-constrained} selection after a single traversal, prioritizing class-separating, non-redundant patterns.

\medskip
\noindent\textbf{Versus multi-phase discriminative miners (DisFPGC \cite{mta-bp}).}
Multi-phase schedules repeatedly enumerate with varying thresholds and only later integrate supervision, which causes redundant work and weak coupling.
DS-Span mines once at a fixed $\delta$, prunes exploration early via eligibility caps, and selects features with an explicit coverage constraint, producing leaner and more predictive feature sets with lower runtime.

\medskip
\noindent\textbf{Versus embedding-first methods.}
Our pipeline is \emph{feature-first}: it yields interpretable substructures with quantified discriminative value.
Embeddings are a thin, optional layer on top of these features, making improvements \emph{attributable} to the mined patterns rather than model capacity.

\section{Implementation \& Experimental Results}
\label{experiments}
\label{Chap_5}

This section distils the evaluation of DS-Span into four viewpoints: (i) accuracy versus existing graph-classification methods, (ii) how many subgraphs are mined (iii) how long mining takes and (iv) how alternative embedding choices affect class separability.

\subsection{Datasets, Protocol, and Hardware}
\label{data_and_exp_setup}
We evaluate on the TU Dataset benchmarks \cite{tudataset}: \textbf{D\&D} \cite{dd}, \textbf{ENZYMES} \cite{enzymes}, \textbf{Proteins} \cite{MASModernApproachToDAI}, \textbf{MUTAG} \cite{mutag}, \textbf{PTC} \cite{ptc}, \textbf{NCI1} and \textbf{NCI109} \cite{nci1}, and \textbf{Reddit-M-5k} \cite{reddit-multi-5k}. For each dataset we regenerate depth-first search frequency caches, run ten repetitions of stratified 10-fold cross-validation with the supervised coverage cap in Section~\ref{subsec:filtering}, and log the number of retained subgraphs per fold. The CBOW-style embedding model uses five epochs, learning rate $\alpha=1.0$, and 64-dimensional representations. All experiments, including the reproduction of \textbf{DisFPGC} \cite{mta-bp}, run on an AMD Ryzen~5~5600G CPU with 32~GB RAM and an RTX~1060 GPU (Python~3.12.3).

Table~\ref{tbl:results} already consolidates the requested accuracy comparison: DS-Span leads on D\&D, ENZYMES, MUTAG, NCI109, Proteins, and PTC, ties Reddit-M-5k, and yields only NCI1 to DisFPGC. To complement those numbers we compare feature budgets and mining costs, then describe how different embedding choices behave.

\subsection{Feature Budget and Mining Cost}
\label{subsec:features_runtime}
\begin{table}[htb!]
\centering
\caption{Average feature counts for DS-Span and DisFPGC.}
\label{tbl:feature_counts}
\begin{tabular}{l|c|c}
\toprule
\textbf{Dataset} & \textbf{Avg. DS-Span features} & \textbf{Avg. DisFPGC features} \\
\midrule
D\&D      & 9.14  & 231.00  \\
Enzymes   & 3.57  & 179.50  \\
Mutag     & 20.02 & 220.60  \\
NCI1      & 12.92 & 283.60  \\
NCI109    & 11.55 & 262.60  \\
Proteins  & 18.99 & 607.60  \\
PTC       & 15.07 & 180.10  \\
\bottomrule
\end{tabular}
\end{table}

\begin{table}[htb!]
\centering
\caption{Average mining times (seconds) for DS-Span and DisFPGC.}
\label{tbl:mining_runtime}
\begin{tabular}{l|c|c}
\toprule
\textbf{Dataset} & \textbf{DS-Span mining} & \textbf{DisFPGC mining} \\
\midrule
D\&D      & 54.61   & 709.80 \\
Enzymes   & 807.23  & 1144.80 \\
Mutag     & 201.24  & 4178.00 \\
NCI1      & 110.01  & 422.30 \\
NCI109    & 96.37   & 362.10 \\
Proteins  & 1455.33 & 7448.20 \\
PTC       & 11.20   & 148.40 \\
\bottomrule
\end{tabular}
\end{table}

Tables~\ref{tbl:feature_counts} and \ref{tbl:mining_runtime} show that DS-Span typically retains fewer than twenty subgraphs and completes mining in seconds, whereas DisFPGC produces hundreds of features and still spends $7$--$265\times$ longer because it repeatedly re-enumerates candidates before applying supervision. Filtering and embedding remain sub-second on all datasets and are omitted for clarity.

\subsection{Embedding Quality and Visual Evidence}
\label{visual}
\begin{figure}[htb!]
     \centering
     \begin{subfigure}[b]{0.32\textwidth}
         \centering
\includegraphics[width=\textwidth]{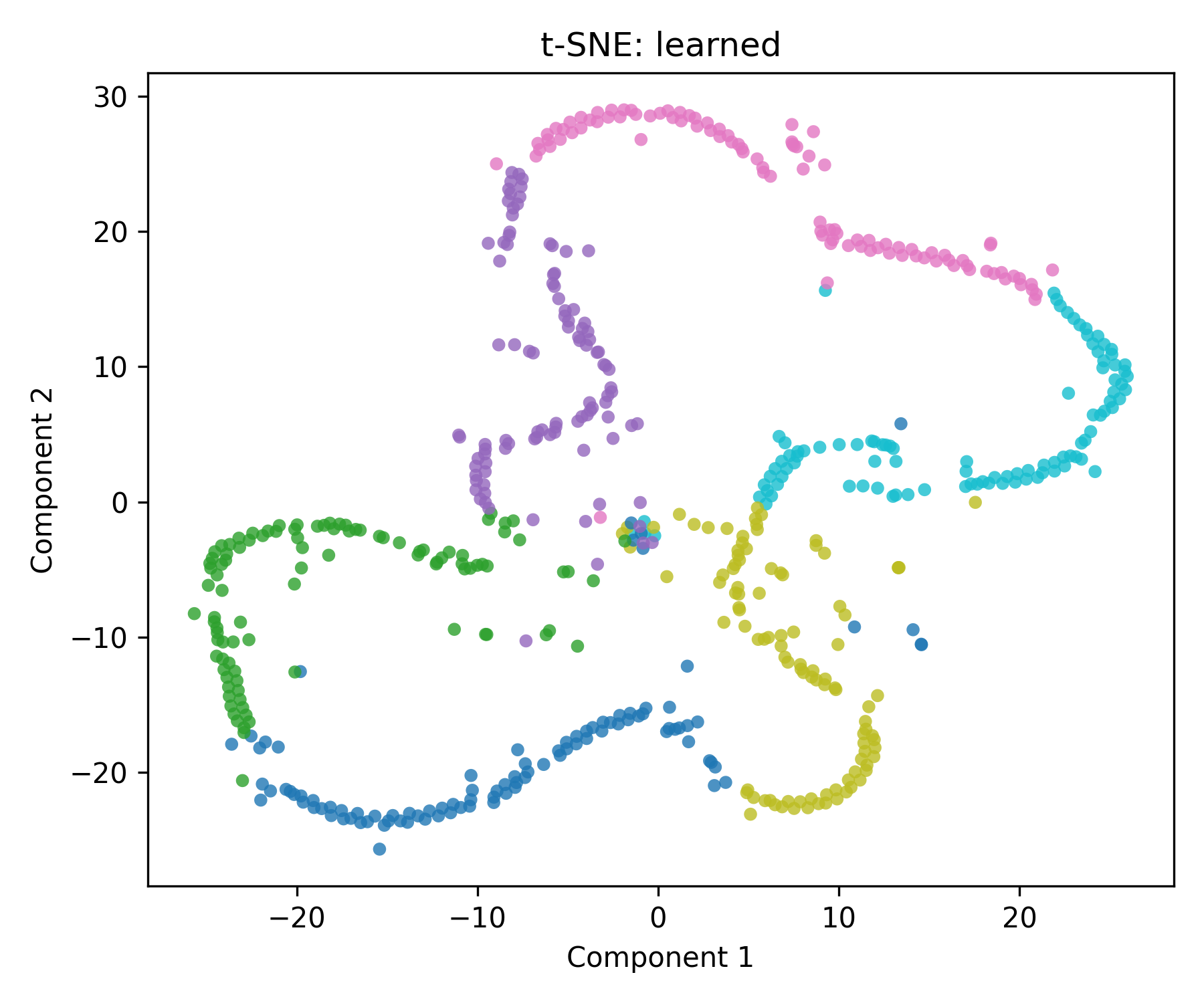}
         \caption{DS-Span}
         \label{fig:tsne-learned}
     \end{subfigure}
     \hfill
     \begin{subfigure}[b]{0.32\textwidth}
         \centering
\includegraphics[width=\textwidth]{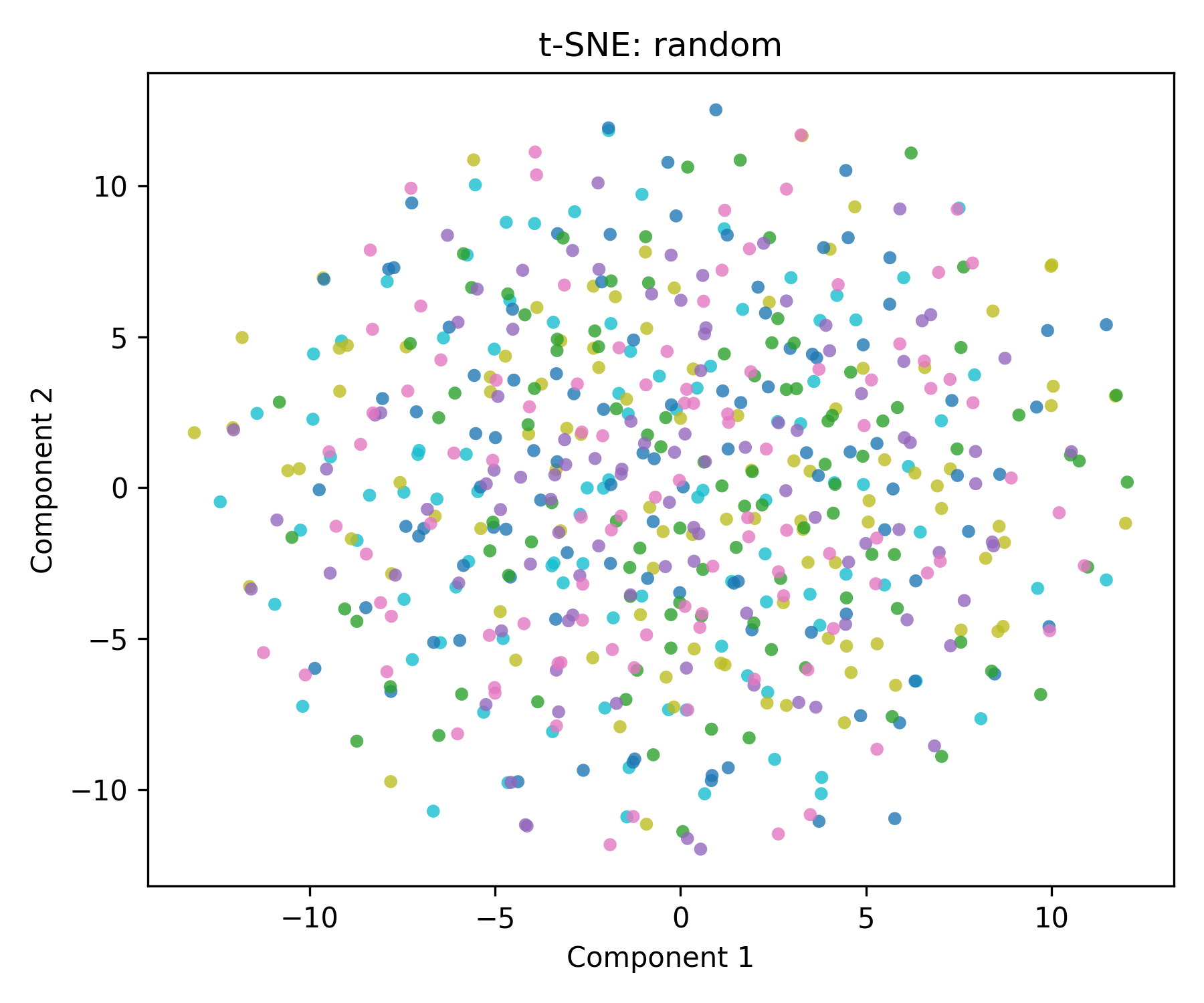}
         \caption{Random baseline}
         \label{fig:tsne-random}
     \end{subfigure}
     \hfill
     \begin{subfigure}[b]{0.32\textwidth}
         \centering
\includegraphics[width=\textwidth]{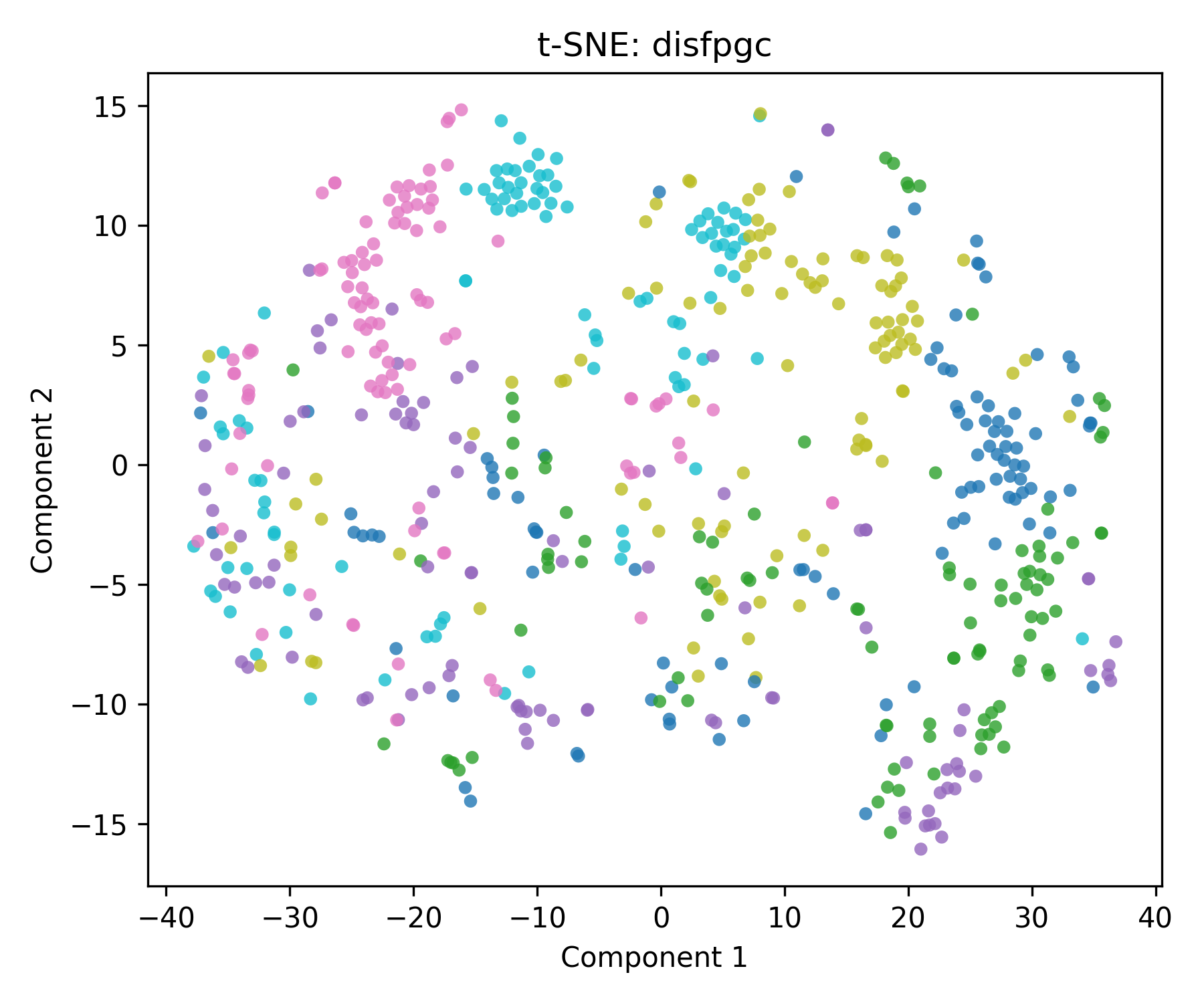}
         \caption{DisFPGC reproduction}
         \label{fig:tsne-disfpgc}
     \end{subfigure}
        \caption{t-SNE visualisations for the D\&D dataset comparing DS-Span embeddings, Gaussian noise, and the reproduced DisFPGC embeddings.}
        \label{fig:tsne-dd}
\end{figure}

To compare embedding choices we visualise DS-Span, a Gaussian random baseline, random-feature embeddings (binary indicators for a random subset of mined subgraphs), their trained counterpart, and the reproduced DisFPGC embeddings. Figure~\ref{fig:tsne-dd} illustrates the D\&D case: DS-Span produces clearly separated clusters despite using only nine patterns, DisFPGC forms worse clusters and requires roughly twenty-five times as many subgraphs, and the random variants collapse into an indistinguishable cloud. Other datasets exhibit the same qualitative behaviour, underscoring that DS-Span's discriminative mining drives its accuracy advantage.

\begin{table}[htb!]
\small
\centering
\definecolor{gold}{RGB}{255, 215, 0}
\definecolor{green}{RGB}{0,200,0}
\caption{Comparison of Methods on Datasets}
\label{tbl:results}
\begin{tabularx}{\textwidth}{p{1.8cm}    % Column 1 width (left-aligned, adjusts automatically)
    |p{0.9cm} % Column 2 width
    |p{1.6cm} % Column 3 width
    |p{1.2cm} % Column 4 width
    |p{1.2cm} % Column 5 width
    |p{1cm} % Column 6 width
    |p{1.3cm} % Column 7 width
    |p{1.45cm} % Column 8 width
    |p{0.8cm} % Column 9 width
}
\toprule
\textbf{Method} & \textbf{D\&D} & \textbf{Enzymes} & \textbf{Reddit-M-5k} & \textbf{Mutag} & \textbf{NCI1} & \textbf{NCI109} & \textbf{Proteins} & \textbf{PTC} \\ 
\midrule
WL-1              & 78.50 \newline \small{(0.44)}   & 50.03 \newline \small{(2.04)}  & -- \newline \small{--}  & 75.35 \newline \small{(2.44)}  & 84.19 \newline \small{(0.41)}  & 84.33 \newline \small{(0.26)}  & 73.03 \newline \small{(0.33)}  & 60.72 \newline \small{(1.80)}  \\ 
\hline
WL-OA             & 78.64 \newline \small{(0.48)}   & 57.02 \newline \small{(1.24)}  & -- \newline \small{--}  & 82.71 \newline \small{(2.34)}  & 85.06 \newline \small{(0.31)}  & 85.06 \newline \small{(0.42)}  & 73.10 \newline \small{(0.81)}  & 60.74 \newline \small{(1.72)}  \\ 
\hline
SP               & 78.47 \newline \small{(0.61)}   & 39.87 \newline \small{(2.01)}  & -- \newline \small{--}  & 82.24 \newline \small{(1.80)}  & 73.97 \newline \small{(0.41)}  & 73.00 \newline \small{(0.28)}  & 75.53 \newline \small{(0.59)}  & 59.36 \newline \small{(1.79)}  \\ 
\hline
$rLap_{BGRL}$       & 74.80 \newline \small{(10.17)}   & 81.50 \newline \small{(5.39)}   & -- \newline \small{--}  & --                & 84.34 \newline \small{(4.03)}  & --                & --                & --                \\ 
\hline
$rLap_{GCL}$    & 70.90 \newline \small{(5.01)}    & \cellcolor{gold} \underline{87.50} \newline \small{(9.86)}   & -- \newline \small{--}  & --                & 75.27 \newline \small{(3.34)}  & --                & --                & --                \\ 
\hline
node2vec         & -- \newline \small{--}                 & 72.63 \newline \small{(10.20)} & -- \newline \small{--}  & 52.68 \newline \small{(1.56)}  & 57.49 \newline \small{(3.57)}  & 58.85 \newline \small{(8.00)}  & -- \newline \small{--}                & -- \newline \small{--}                \\ 
\hline
sub2vec          & -- \newline \small{--}                 & 61.05 \newline \small{(15.79)} & -- \newline \small{--}  & 50.67 \newline \small{(1.50)}  & 53.03 \newline \small{(5.55)}  & 59.99 \newline \small{(6.38)}  & -- \newline \small{--}                & -- \newline \small{--}                \\ 
\hline
graph2vec        & 58.64 \newline \small{(0.01)}                 & 44.33  \newline \small{(0.09)}  & -- \newline \small{--}  & 83.15  \newline \small{(9.25)}  & 73.22  \newline \small{(1.81)}  & 74.26  \newline \small{(1.47)}  & 73.30  \newline \small{(2.05)}                & 60.17 \newline \small{(6.86)}                \\ 
\hline
HGP-SL            & 80.96 \newline \small{(1.26)}   & 68.79 \newline \small{(2.11)}   & -- \newline \small{--}  & -- \newline \small{--}  & 78.45 \newline \small{(0.77)}  & 80.67 \newline \small{(1.16)} & \cellcolor{gold} \underline{84.91} \newline \small{(1.62)} & -- \newline \small{--} \\
\hline
GraphSage         & 72.90 \newline \small{(2.00)}    & 58.20 \newline \small{(6.00)}   & 50.0 \newline \small{(1.3)}  & 79.80 \newline \small{(13.90)}  & 76.00 \newline \small{(1.80)}  & -- \newline \small{--} & 73.00 \newline \small{(4.50)} & -- \newline \small{--} \\
\hline
\footnotesize{$SAGPool_g$}         & 76.19 \newline \small{(0.94)}  & -- \newline \small{--}   & -- \newline \small{--}  & 90.42 \newline \small{(7.78)}  & 74.18 \newline \small{(1.20)}  & 74.06 \newline \small{(0.78)}  & 70.04 \newline \small{(1.47)} & -- \newline \small{--} \\
\hline
DiffPool          & 75.00 \newline \small{(3.50)}   & 59.50 \newline \small{(5.60)}  & 53.8 \newline \small{(1.4)}  & 77.60 \newline \small{2.70}  & 76.90 \newline \small{(1.90)}  & -- \newline \small{--}  & 72.70 \newline \small{3.80}  & 73.70 \newline \small{(3.50)} \\
\hline
$GIN_{\epsilon-JK}$   & -- \newline \small{--}                 & 39.30 \newline \small{(1.60)}                & 57.0 \newline \small{(1.7)}                & -- \newline \small{--}                & 78.30 \newline \small{(0.30)}                & -- \newline \small{--}                & 72.20 \newline \small{(0.70)}                & -- \newline \small{--}                \\ 
\hline
GCN               & -- \newline \small{--}       & -- \newline \small{--}   & -- \newline \small{--}  & 85.60 \newline \small{(5.80)}  & 80.20 \newline \small{(2.00)}  & -- \newline \small{--}  & 76.00 \newline \small{(3.20)}  & 64.20 \newline \small{(4.30)} \\
\hline
GCKN & -- \newline \small{--}       & -- \newline \small{--}       &-- \newline \small{--}        & \cellcolor{gold} \underline{97.20} \newline \small{(2.80)} & 83.90 \newline \small{(1.20)}& -- \newline \small{--}        & 75.90 \newline \small{(3.20)} & 69.40 \newline \small{(3.50)}\\
\hline
U2GNN            & \cellcolor{gold} \underline{95.67} \newline \small{(1.89)}  & -- \newline \small{--}   & -- \newline \small{--}  & 88.47 \newline \small{(7.13)}  & -- \newline \small{--}  & -- \newline \small{--}  & 80.01 \newline \small{(3.21)} & \cellcolor{gold} \underline{91.81} \newline \small{(6.61)} \\
\hline
GIU-Net           & -- \newline \small{--}       & 70.00 \newline \small{--}   & -- \newline \small{--}  & 95.70 \newline \small{--}  & 80.20 \newline \small{--}  & 77.00 \newline \small{--}  & 77.60 \newline \small{--}  & 85.70 \newline \small{--} \\
\hline
GE-FSG           & 91.69 \newline \small{(0.02)}   & 49.33 \newline \small{(0.07)}  & -- \newline \small{--}  & 84.74 \newline \small{(0.07)}  & 84.36 \newline \small{(0.02)}  & 85.59 \newline \small{(0.01)}  & 81.79 \newline \small{(0.04)}  & 62.57 \newline \small{(0.09)}  \\ 
\hline
DisFPGC          & 93.46 \newline \small{(0.54)}   & 59.82 \newline \small{(0.16)}  & \cellcolor{gold} \underline{93.10} \newline \small{(1.52)}  & 97.00 \newline \small{(0.53)}   & \cellcolor{green} \textbf{94.92} \newline \small{(0.21)}  & \cellcolor{gold} \underline{97.40} \newline \small{(0.12)}   & 83.44 \newline \small{(0.67)}  & 84.94 \newline \small{(0.50)}  \\ 
\hline
\textbf{DS-SPAN}               & \cellcolor{green} \textbf{96.12} \newline \small{(0.29)}       & \cellcolor{green}\textbf{93.80} \newline \small{(0.59)}   & \cellcolor{green} \textbf{93.19} \newline \small{(0.25)}  & \cellcolor{green} \textbf{99.67} \newline \small{(0.27)}  & \cellcolor{gold} \underline{89.44} \newline \small{(0.31)}  & \cellcolor{green} \textbf{99.43} \newline \small{(0.08)}  & \cellcolor{green} \textbf{93.89} \newline \small{(0.59)}  & \cellcolor{green} \textbf{95.03} \newline \small{(0.83)} \\

\bottomrule
\end{tabularx}
\end{table}

\newpage
\section{Conclusion}
\label{Conc}
In this work, we proposed DS-Span, a computationally efficient approach for generating whole-graph embeddings by leveraging a single-phase discriminative subgraph mining technique. Our method addresses the challenges posed by traditional multi-phase approaches, such as high computational cost and redundancy, by dynamically shrinking the search space and employing a robust filtering mechanism to identify highly discriminative subgraphs. These subgraphs form the basis for embeddings that preserve critical structural and class-separating features. The contributions of our work are multifaceted. First, we developed a single-phase subgraph mining algorithm that eliminates iterative overhead and significantly enhances scalability. Second, we introduced a supervised graph embedding methodology that outperforms existing state-of-the-art methods in classification tasks, as demonstrated across diverse benchmark datasets. Lastly, we showed the practical applicability of our approach in critical domains, such as drug discovery, by enabling precise and efficient classification of molecular graphs, thereby reducing resource-intensive experimental validation. The results indicate that our method attains greater accuracy than baseline methods and state-of-the-art techniques in most datasets, with a very low standard deviation. This reflects the robustness and stability of our work. Specifically, it indicates our approach effectively discriminates between classes and generates high-quality whole-graph embeddings useful for downstream graph analytics. As our approach is computationally efficient and highly optimized, it requires fewer resources and processing time than its predecessors with similar techniques. These findings underline the effectiveness of our method in graph classification and suggest its potential for practical applications. In future work, we aim to explore unsupervised filtering mechanisms to enhance the generalizability of our method and investigate dataset-specific hyperparameter tuning to optimize performance across diverse graph datasets even further.

\bibliographystyle{splncs04}

\bibliography{samplepaper}

\begin{thebibliography}{10}
\providecommand{\url}[1]{\texttt{#1}}
\providecommand{\urlprefix}{URL }
\providecommand{\doi}[1]{https://doi.org/#1}

\bibitem{mta-bp}
Alam, M.T., Ahmed, C.F., Samiullah, M., Leung, C.K.: Discriminating frequent pattern based supervised graph embedding for classification. In: PAKDD (2021)

\bibitem{enzymes}
Borgwardt, K.M., Ong, C.S., Schonauer, S., Vishwanathan, S.V.N., Smola, A.J., Kriegel, H.p.: {Protein function prediction via graph kernels}. Bioinformatics  \textbf{21}(Suppl 1),  i47--i56 (6 2005). \doi{10.1093/bioinformatics/bti1007}, \url{https://doi.org/10.1093/bioinformatics/bti1007}

\bibitem{MASModernApproachToDAI}
Borgwardt, K.M., O.C.S.S.V.S.S.A.K.H.: Protein function prediction via graph kernels  \textbf{3} (10 2005)

\bibitem{mutag}
Debnath, A.K., Lopez~de Compadre, R.L., Debnath, G., Shusterman, A.J., Hansch, C.: Structure-activity relationship of mutagenic aromatic and heteroaromatic nitro compounds. correlation with molecular orbital energies and hydrophobicity. Journal of Medicinal Chemistry  \textbf{34}(2),  786--797 (1991). \doi{10.1021/jm00106a046}, \url{https://doi.org/10.1021/jm00106a046}

\bibitem{dd}
Dobson, P.D., Doig, A.J.: {Distinguishing Enzyme Structures from Non-enzymes Without Alignments}. Journal of Molecular Biology  \textbf{330}(4),  771--783 (7 2003). \doi{10.1016/s0022-2836(03)00628-4}, \url{https://doi.org/10.1016/s0022-2836(03)00628-4}

\bibitem{tudataset}
Morris, C., Kriege, N.M., Bause, F., Kersting, K., Mutzel, P., Neumann, M.: Tudataset: A collection of benchmark datasets for learning with graphs. In: ICML 2020 Workshop on Graph Representation Learning and Beyond (2020)

\bibitem{ptc}
Toivonen, H., Srinivasan, A., King, R.D., Kramer, S., Helma, C.: {Statistical evaluation of the Predictive Toxicology Challenge 2000–2001}. Bioinformatics  \textbf{19}(10),  1183--1193 (07 2003). \doi{10.1093/bioinformatics/btg130}, \url{https://doi.org/10.1093/bioinformatics/btg130}

\bibitem{nci1}
Wale, N., Watson, I., Karypis, G.: Comparison of descriptor spaces for chemical compound retrieval and classification. Knowledge and Information Systems  \textbf{14}(3),  347--375 (Mar 2008). \doi{10.1007/s10115-007-0103-5}

\bibitem{gspan}
Yan, X., Han, J.: gspan: graph-based substructure pattern mining. In: 2002 IEEE International Conference on Data Mining, 2002. Proceedings. pp. 721--724 (2002). \doi{10.1109/ICDM.2002.1184038}

\bibitem{reddit-multi-5k}
Yanardag, P., Vishwanathan, S.: {Deep Graph Kernels}. KDD '15: Proceedings of the 21th ACM SIGKDD International Conference on Knowledge Discovery and Data Mining pp. 1365--1374 (8 2015). \doi{10.1145/2783258.2783417}, \url{https://doi.org/10.1145/2783258.2783417}

\end{thebibliography}

\end{document}